\pdfoutput=1

\documentclass[11pt]{article}

\usepackage[final]{acl}

\usepackage{times}
\usepackage{latexsym}

\usepackage[T1]{fontenc}

\usepackage[utf8]{inputenc}

\usepackage{microtype}

\usepackage{inconsolata}

\usepackage{graphicx}
\usepackage{amsmath, amssymb}
\usepackage{booktabs}
\usepackage{algorithm}
\usepackage{algpseudocode}
\usepackage{multirow}

\title{GiFT: Gibbs Fine-Tuning for Code Generation}

\author{{\bf Haochen Li}\ \ \ \ 
{\bf Wanjin  Feng}\ \ \ \
    {\bf Xin Zhou\thanks{\ \ Corresponding author}}\ \ \ \ 
    {\bf Zhiqi Shen} \ \ \ \ \\
     Nanyang Technological University, Singapore \\
     \texttt{\{haochen003, xin.zhou, zqshen\}@ntu.edu.sg \ \ \texttt{wanjin\_feng@outlook.com}
     }}

\begin{document}
\maketitle
\begin{abstract}
Training Large Language Models (LLMs) with synthetic data is a prevalent practice in code generation. A key approach is self-training, where LLMs are iteratively trained on self-generated correct code snippets. In this case, the self-generated codes are drawn from a conditional distribution, conditioned on a specific seed description. However, the seed description is not the only valid representation that aligns with its intended meaning.  
With all valid descriptions and codes forming a joint space, codes drawn from the conditional distribution would lead to an underrepresentation of the full description-code space.
As such, we propose \textbf{Gi}bbs \textbf{F}ine-\textbf{T}uning (GiFT), a novel self-training method inspired by Gibbs sampling. GiFT allows self-generated data to be drawn from the marginal distribution of the joint space, thereby mitigating the biases inherent in conditional sampling.
We provide a theoretical analysis demonstrating the potential benefits of fine-tuning LLMs with code derived from the marginal distribution. 
Furthermore, we propose a perplexity-based code selection method to mitigate the imbalanced long-tail distribution of the self-generated codes.
Empirical evaluation of two LLMs across four datasets demonstrates that GiFT achieves superior performance, particularly on more challenging benchmarks.
Source code is available at \url{https://github.com/Alex-HaochenLi/GiFT}.

\end{abstract}

\section{Introduction}
Code generation, the automated synthesis of code snippets from natural language specifications, significantly enhances software development productivity~\cite{archcode, self-planning}. Recent advances in large language models (LLMs), trained on massive web-derived code and text corpora, exhibit notable capabilities for code understanding and generation~\cite{codegensurvey,wang2024openhands,liu2024large}. While scaling training data is beneficial, high-quality data has been found to play a more important role in boosting LLM performance~\cite{magicoder, selfcodealign}.

However, curating large-scale, high-quality datasets by manual annotation is challenging due to substantial costs. Consequently, researchers switch gears to synthetic data. Two principal approaches to generate data are: a) knowledge distillation from stronger LLMs~\cite{magicoder,wavecoder,wizardcoder}, and b) self-training, wherein models generate their own training data~\cite{rft,star,selfcodealign}. The prerequisite of stronger LLMs for distillation restricts its general applicability, thus many works focus more on self-training, as we studied in this paper. 

\begin{figure}[t]
\centering
\includegraphics[width=0.8\columnwidth, trim=0 25 0 50, clip]{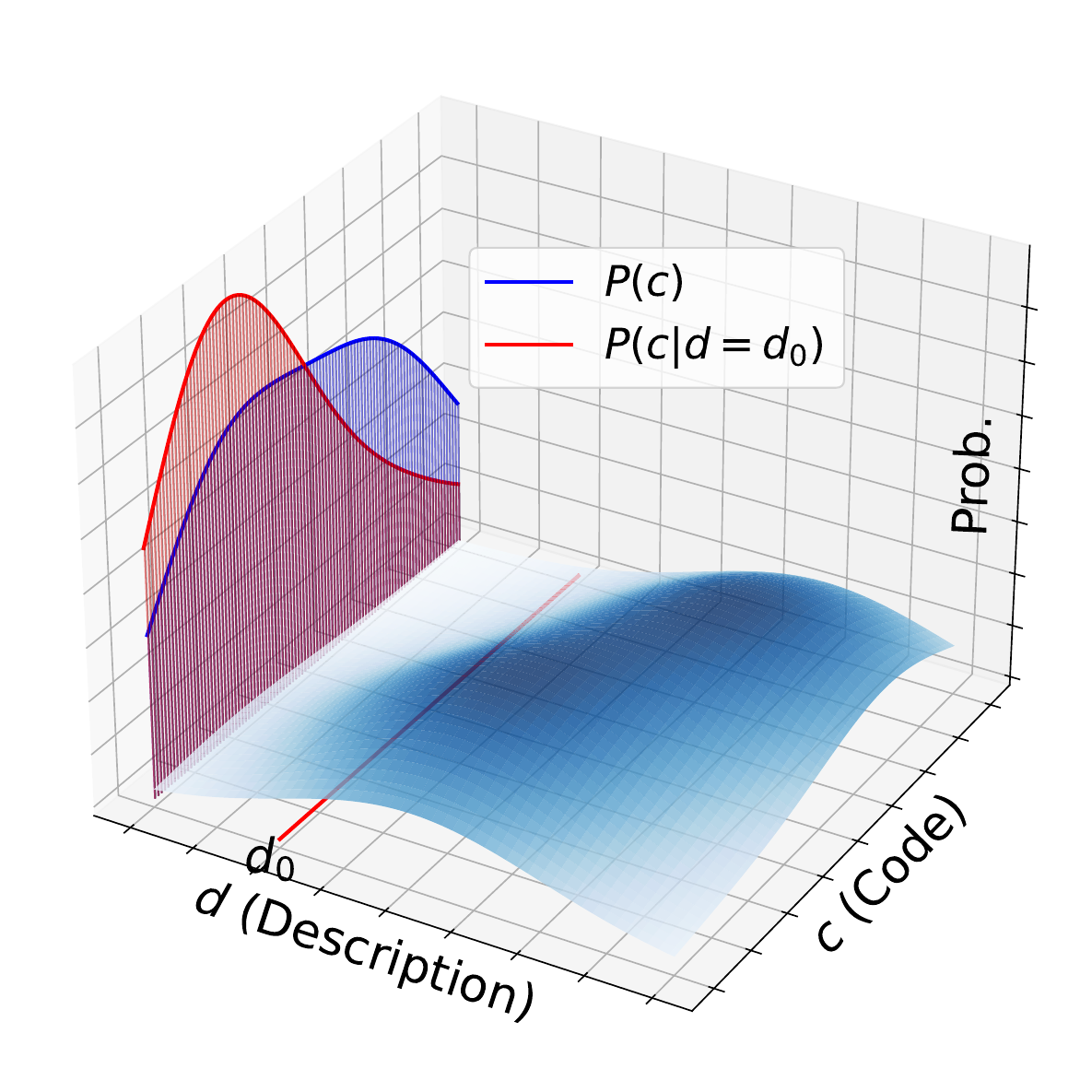}
\caption{For the intention of $d_0$, the set of all valid descriptions and codes forms a space. The distribution gap between conditional distribution (\textcolor{red}{Red}) and marginal distribution (\textcolor{blue}{Blue}) indicates the bias introduced when fine-tuned LLMs with codes conditional on $d_0$, as some codes are rarely sampled.}
\label{fig1}
\vspace{-5pt}
\end{figure}

\begin{figure*}[t]
\centering
\includegraphics[width=1.99\columnwidth, trim=0 450 300 0, clip]{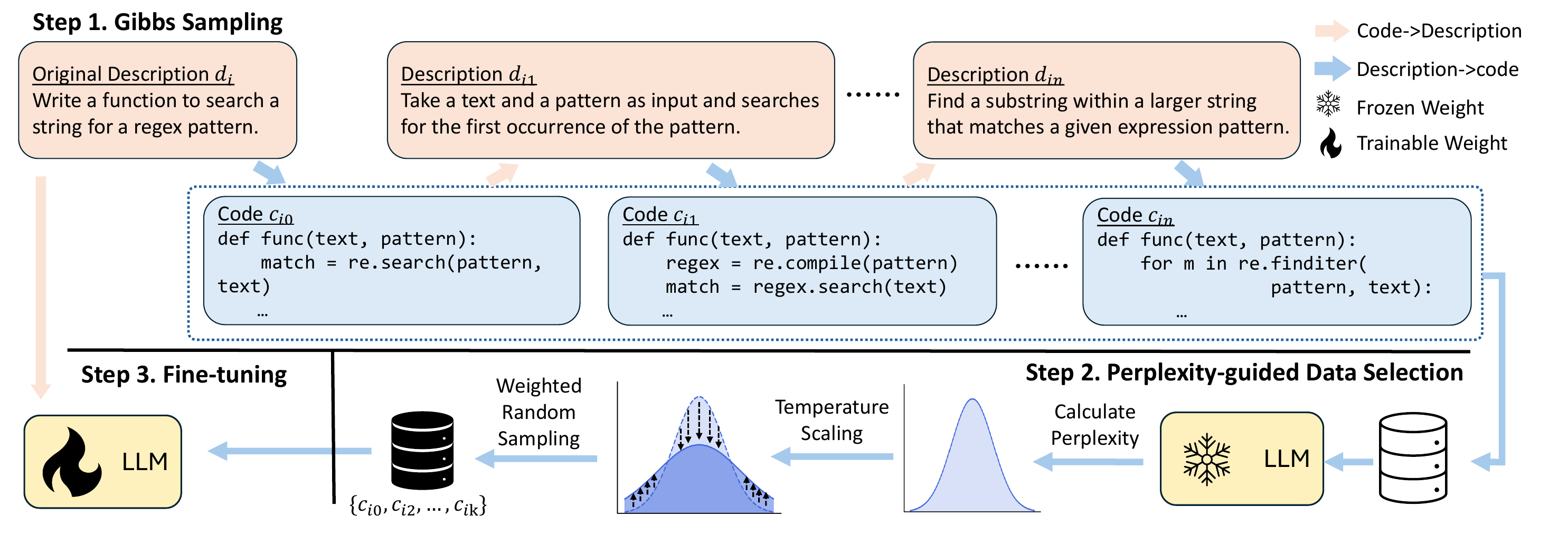}
\caption{Overview of GiFT. For each description $d_i$ in the seed dataset, we first translate it between descriptions and codes iteratively to draw codes from the marginal distribution based on the intention of $d_i$. Then, we calculate the perplexity of each generated code and employ weighted random sampling to select codes with codes from the tail being more likely to be selected for fine-tuning. Finally, all selected codes are paired with $d_i$ for fine-tuning. The example shown in this figure is taken from MBPP-sanitized/6.}
\label{overview}
\vspace{-5pt}
\end{figure*}

The self-training process utilizes a seed dataset, denoted as $\{d_i\}_{i=1}^{N}$. For each description $d_i$, an LLM generates multiple candidate code snippets, which are evaluated against test cases. Snippets passing all tests are used to fine-tune the same LLM. This process iterates until performance plateaus or degrades.
However, the natural language description $d_i$ represents only one possible articulation of the underlying intention. Consider the goal of matching a pattern within a string. This underlying goal—what we refer to as the ``intention''—can be expressed through various descriptions such as ``\textit{Write a function to search a string for a regex pattern}'' or ``\textit{Find a substring within a larger string that matches a given regular expression pattern}''. Considering the set of all valid descriptions and corresponding code implementations that satisfy the underlying intention of $d_i$ as a description-code space, generating code exclusively from $d_i$ can be viewed as sampling from the conditional distribution $P(c|d_i)$. We argue that this approach is suboptimal. Instead, we propose that sampling from the marginal distribution of codes within the joint description-code space, denoted as $P(c)$, would yield superior results. Figure~\ref{fig1} illustrates the potential benefit of sampling from the marginal distribution compared to the conditional distribution.  The figure highlights the possibility of oversampling certain code implementations and undersampling others when relying solely on the conditional probability $P(c|d_i)$.

In this paper, we first theoretically justify the benefit of fine-tuning LLMs with samples drawn from the marginal distribution, by showing that an additional expectation of loss is implicitly taken to reduce the bias introduced by samples from the conditional distribution.
Direct sampling from the marginal distribution is intractable in practice. We gain inspiration from Gibbs sampling~\cite{gibbssampling}, an MCMC algorithm,
that iteratively samples each variable from its conditional distribution while keeping the others fixed, gradually approximating the joint distribution.
Simulating the Gibbs sampling in the context of code generation, we propose \textbf{Gi}bbs \textbf{F}ine-\textbf{T}uning (GiFT). 
From a seed description, code is generated and then summarized into a new description, used for subsequent code generation. 
This process is repeated to get a set of self-generated description-code pairs, which could be considered drawn from the joint distribution. The code components in pairs can be considered drawn from the marginal distribution
\footnote{In this paper, we primarily study the impact of using code samples from either conditional or marginal distributions on fine-tuning. The incorporation of self-generated descriptions is discussed in Section~\ref{sec:exp}.}.

Self-generated code often suffers from data imbalances detrimental to LLM fine-tuning. One source of imbalance is the varying number of generated codes for descriptions of differing difficulty. We address this by selecting a fixed number of codes per description.
The other more fundamental imbalance arises from the long-tailed nature of the marginal distribution from which code is sampled~\cite{ding2024mitigating,dohmatob2024tale}. High-probability (head) codes are over-sampled, while low-probability (tail) codes are under-sampled. This disparity can lead to model collapse during iterative self-training~\cite{collapse}. To address this, we use perplexity as a proxy for the likelihood of being sampled: higher perplexity indicates rarer, tail-distributed code. During training, we select more high-perplexity codes from the tail. Figure~\ref{overview} illustrates the GiFT overview.

We evaluate GiFT on DeepSeek-Coder-6.7B \cite{deepseek-coder} and CodeLlama-7B~\cite{codellama} over APPS+ (Introductory-level and Interview-level)~\cite{apps+}, MBPP+~\cite{mbpp+}, and CodeInsight~\cite{codeinsight} datasets. Experimental results demonstrate the superiority of drawing codes from the marginal distribution instead of the conditional distribution (+1.2\% on MBPP+, +2.3\% on CodeInsight, +9.8\% on APPS+ dataset), and perplexity-guided data selection benefits self-training over iterations.

\section{Related Work}
We classify related works into distillation and self-training based on whether the synthetic data is generated by stronger LLMs or the LLM undergoing training itself.
\paragraph{Distillation}
Code Alpaca~\cite{codealpaca}, similar to Self-Instruct~\cite{selfinstruct}, leverages the in-context learning ability of ChatGPT to generate new description-code pairs. WizardCoder~\cite{wizardcoder} prompts ChatGPT with five tailored heuristics to improve the difficulty of existing descriptions in Code Alpaca. Magicoder~\cite{magicoder} and WaveCoder~\cite{wavecoder} highlight the importance of data diversity and quality by prompting ChatGPT to create new pairs based on open-sourced codes on the web instead of LLM-generated Code Alpaca. 
MathGenie~\cite{mathgenie} improves from the solution side where it augments the solutions by prompting an external LLM with heuristics and then back-translates augmented solutions into math problems in order to create new problems. However, stronger LLMs are not always available, which limits the generalizability of distillation methods.

\paragraph{Self-training}
Self-training refers to making LLMs learn from their own outputs based on a set of seed descriptions. Self-training approaches can be categorized into two directions based on whether additional data is synthesized on the description side or the code side. On the description side, Instruction Backtranslation~\cite{instruction-backtranslation} and InverseCoder~\cite{inversecoder} ask an LLM to generate synthesized descriptions for unlabeled codes for instruction tuning.
On the code side, Self-Taught Reasoner (STaR)~\cite{star} is a pioneering work that generates a single rationale for each reasoning problem. LMSI~\cite{lmsi} and Rejection Fine-tuning (RFT)~\cite{rft} enhance STaR by generating multiple rationales per problem. While STaR, LMSI, and RFT rely on the ground truth answer to filter out incorrect rationales, SelfCodeAlign~\cite{selfcodealign} additionally asks LLMs to generate test cases for synthesized codes to conduct self-validation. Reinforced Self-Training (ReST)~\cite{rest} and ReST$^{EM}$~\cite{restem} expand the RFT process into an iterative one, where the generate-then-fine-tune process is repeated multiple times until no further improvement is observed.

Though there are intermediate descriptions generated in GiFT, we do not use those intermediate descriptions for fine-tuning, as GiFT is mainly proposed to improve the data quality on the code side.
GiFT is orthogonal to the self-training methods on the code side as each synthetic description can benefit from higher-quality codes generated in GiFT. Besides, GiFT is beneficial under the distillation setting. We empirically demonstrate the effectiveness of data from GiFT in Section~\ref{sec:exp}.

\section{Gibbs Fine-Tuning}
\paragraph{Preliminaries}
We first introduce how iterative self-training works. Given a seed dataset $\mathcal{D}=\{d_i\}_{i=1}^N$, an LLM $\mathcal{M}$ is used to generate $n$ code snippets for each $d_i$:
\begin{equation}
    \{c_{ij}\}_{j=1}^n \sim P_{\mathcal{M}}(c|d_i)
\end{equation}
Then correct codes that pass all the test cases are selected as $\mathcal{C}$ for supervised fine-tuning (SFT). The SFT loss $\mathcal{L}$ for $d_i$ could be written as:
\begin{align}
\label{eq:sft}
    \mathcal{L}(d_i^*) &= - \mathbb{E}_{\mathcal{C}\sim P_{\mathcal{M}}(c|d_i^*)} \log P_{\mathcal{M}}(c|d_i) \nonumber \\
    & = - \sum_{c\in \mathcal{C}} P_{\mathcal{M}}(c|d_i^*) \log P_{\mathcal{M}}(c|d_i)
\end{align}
Here $d_i^*$ refers to the description source that $c_{ij}$ is generated from. In this case, $d_i^*=d_i$. The SFT loss is calculated over the seed dataset $\mathcal{D}$ to update $\mathcal{M}$, and the updated $\mathcal{M}$ will be used to generate codes in the next iteration. This process is repeated until no further improvement in performance is observed.

\paragraph{Theoretical Insight}
The problem in the code generation process is that all the self-generated codes are drawn from a conditional distribution $c_{ij}\sim P(c|d_i)$ instead of the joint space of descriptions and codes based on the intention behind $d_i$. We argue that it is better to draw codes from the marginal distribution of that space. If we fine-tune LLMs with codes from marginal distribution, we have:
\begin{align}
    & \mathcal{L}_{marg} \nonumber\\
    &= - \mathbb{E}_{c\sim P_c} \log P_{\mathcal{M}}(c|d_i) \nonumber \\
    &= - \sum_{c} P_c(c) \log P_{\mathcal{M}}(c|d_i) \nonumber \\
    &= - \sum_{c}\sum_{d_{ij}} P_{\mathcal{M}}(c|d_{ij})P(d_{ij}) \log P_{\mathcal{M}}(c|d_i) \nonumber \\
    &= - \sum_{d_{ij}}\Big [ \sum_{c} P_{\mathcal{M}}(c|d_{ij}) \log P_{\mathcal{M}}(c|d_i)\Big ] 
 P(d_{ij}) \nonumber \\
    &= - \sum_{d_{ij}} \mathcal{L}(d_{ij}) P(d_{ij}) \label{eq3}\\
    &= - \mathbb{E}_{d_{ij}\sim P_d} \mathcal{L}(d_{ij})
\end{align}
According to the law of total expectation, we could find that $\mathcal{L}_{marg}$ is estimated over all possible descriptions $d_{ij}$ in the joint space, instead of only $d_i$ in Eq.~\ref{eq:sft}. The additional expectation in $\mathcal{L}_{marg}$ reduces the bias of $\mathcal{L}(d_i)$ in learning to generate codes for the intention behind $d_i$.

Besides, we analyze the variance of self-generated codes $c$ from either the marginal distribution or the conditional distribution. According to the law of total variance, we have:
\begin{align}
Var(c) = \mathbb{E}_{d_{ij}}[Var(c|d_{ij})] + Var(\mathbb{E}_{d_{ij}}[c|d_{ij}]) \nonumber
\end{align}
Since $Var(\mathbb{E}_{d_{ij}}[c|d_{ij}]) \ge 0$, the variance of codes drawn from the marginal distribution is greater than or equal to the expected variance of codes drawn from the conditional distribution conditioned on a certain $d_{ij}$ (e.g. the seed description $d_i$). What's more, $Var(\mathbb{E}_{d_{ij}}[c|d_{ij}])$ could become even larger if an LLM is sensitive to input descriptions, which further widens the gap between $Var(c)$ and $\mathbb{E}_{d_{ij}}[Var(c|d_{ij})]$. Here the variance of $c$ reflects the diversity of self-generated codes. More diverse codes are found to benefit LLM fine-tuning~\cite{rft}.

\paragraph{Gibbs Sampling}
Though we have demonstrated that marginal distribution is better than conditional distribution, direct sampling from marginal distribution is not straightforward, as we only have one certain $d_i$ in the seed dataset. We gain inspiration from Gibbs sampling~\cite{gibbssampling}, a Markov chain Monte Carlo algorithm, that is commonly used to approximate joint distributions based on conditional distributions. Take a bivariate distribution as an example. It approximates joint distributions by drawing an instance of one variable conditional on the current value of the other variable, then drawing an instance of the other variable conditional on the new value of the first variable, and repeating this process for several rounds.

In GiFT, we consider the code-to-text translation and text-to-code translation as the conditional sampling process. We keep translating between descriptions and codes to simulate Gibbs sampling. During this process, all the intermediate description-code pairs could be considered as being drawn from the joint distribution.  
If we take all codes from the pairs, those codes can be considered drawn from the marginal distribution of the joint space.
Specifically, for each description $d_i$ in the seed dataset, we start from the description side to generate corresponding codes $c_{i1}$, and then summarize $c_{i1}$ into description $d_{i1}$. We repeat this process for $n$ times. The whole process could be formulated as:
\begin{align}
    c_{i1} = \mathcal{M}(d_i) & \qquad
    d_{i1} = \mathcal{M}(c_{i1}) \nonumber \\ 
    &\vdots  \\
    d_{in-1} = \mathcal{M}(c_{in-1}) & \qquad
    c_{in} = \mathcal{M}(d_{in-1})\nonumber
\end{align}
The prompting templates for code generation and summarization are shown in Appendix~\ref{app:prompt}. To improve the efficiency of the Gibbs sampling, we generate 3 codes in each code generation step but only select one correct code for the following rounds. If none of the 3 codes passes all the test cases, we use the code from the last round for the next code summarization step.

\paragraph{Perplexity-guided Data Selection}
After the Gibbs sampling process, for each $d_i$, we have a set of codes that could be paired with it for fine-tuning. As shown in Eq.~\ref{eq3}, $P(d_{ij})$ plays a pivotal role in the estimation of $\mathcal{L}_{marg}$. In practice, $P(d_{ij})$ is reflected in the selection of codes.

Simply selecting all correct codes is detrimental. On the one hand, we are more likely to sample more codes for easy descriptions in the seed dataset and less for harder ones. Fine-tuning with all codes will bias LLMs towards easy descriptions~\cite{restem}, so we only select $K$ codes for each description. For descriptions with fewer than $K$ codes, we resample existing codes to ensure balance. On the other hand, there is data imbalance in the code set of each $d_i$. According to \citet{ding2024mitigating}, the marginal distribution within each code set is found to follow a long-tail distribution. Employing random sampling to select $K$ codes makes codes from the tail occupy only a small proportion of the training data since they are seldom generated, which bias $\mathcal{L}_{marg}$ towards the head. In iterative self-training, this bias will be exacerbated where the knowledge distribution of the LLM shifts to be more peaked.

We propose to use perplexity~\cite{perplexity} as a measurement to guide data selection.
As we know, LLMs prefer tokens with higher probabilities in each generation step, despite using temperature to flatten the probability distribution. The probability of generating a code $c$ with $l$ tokens given $d$ could be formulated as:
\begin{equation}
    P_{\mathcal{M}}(c|d)=\prod_{t=1}^l P_{\mathcal{M}}(c_t|c_{<t},d) \nonumber
\end{equation}
And the perplexity (ppl) is calculated by:
\begin{equation}
    \mathrm{ppl}(c|d,\mathcal{M}) = \exp \Big( -{\frac{1}{l}}\sum_{t=1}^l \log P_{\mathcal{M}}(c_t|c_{<t},d)\Big) \nonumber
\end{equation}
We could find that the perplexity of $c$ and the probability of generating $c$ have a strong negative correlation. In other words, codes with lower perplexity are more likely to come from the head. Thus, to mitigate imbalance during data selection, we employ weighted random sampling and assign more weights for high perplexity (i.e. tail) codes:
\begin{equation}
w_{ij} = \frac{\exp (\mathrm{ppl}(c_{ij}) / T)}{\sum_{j=1}^{n_i} \exp (\mathrm{ppl}(c_{ij}) / T)} \nonumber
\end{equation}
where $n_i$ is the number of correct codes for $d_i$ and $T$ is the scaling temperature. Finally, the selected codes $\{c_{ij}\}_{j=1}^K$ are paired with $d_i$ for fine-tuning LLMs with the SFT loss.
The workflow of GiFT in each iteration is shown in Appendix~\ref{app:algo} Algorithm~\ref{algo}.

\begin{figure*}[ht]
\centering
\includegraphics[width=1.99\columnwidth, trim=0 0 0 0, clip]{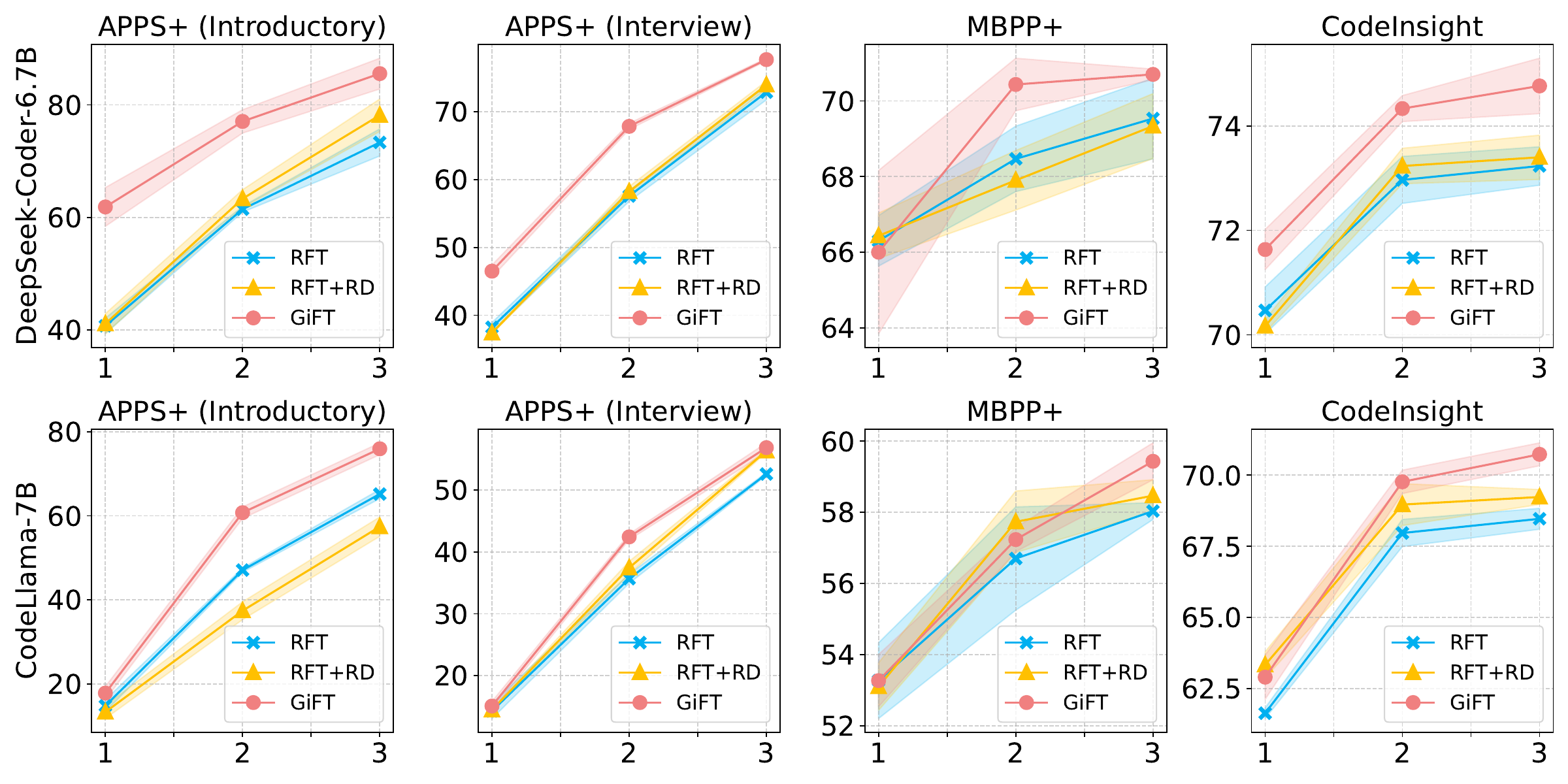}
\caption{Pass@1 (\%) of applying RFT, RFT+RD, and GiFT to Deepseek-Coder-6.7B and CodeLlama-7B on 4 code generation datasets. The x-axis represents the iteration number and the shaded area represents the standard deviation.}
\label{overall_result}
\vspace{-5pt}
\end{figure*}

\section{Experimental Setups}
\paragraph{Datasets}
We evaluate GiFT on three datasets, APPS+~\cite{apps+}, MBPP~\cite{mbpp}, and CodeInsight~\cite{codeinsight}. APPS+ is a sanitized version of APPS~\cite{apps} where wrong descriptions or test cases are removed from the original dataset. For MBPP, we use MBPP-sanitized for training and MBPP+~\cite{mbpp+} for testing. The APPS dataset consists of problems collected from different open-access coding websites, MBPP is full of general programming problems, and CodeInsight is collected from StackOverflow focusing on standard library usage. In this paper, we consider the problems in APPS+ with the difficulty of ``introductory'' and ``interview'' as two independent datasets. All four datasets are written in Python and their statistics are shown in Appendix~\ref{app:setup}. We take the widely adopted Pass@1 as the evaluation metric.

\paragraph{Baselines}
We compare GiFT with two baseline methods. (1) Rejection Fine-Tuning (\textbf{RFT})~\cite{rft} uses rejection sampling that generates multiple codes depending on each seed description. As ReST~\cite{rest} could be considered as iterative RFT, we denote this baseline as RFT in our experiments. (2) In RFT+Rewriting Description (\textbf{RFT+RD}), we first ask LLMs to rewrite the seed description and then apply RFT to both the original description and rewritten descriptions. Though no related works employ this method, we consider it as an alternative way to approximate the marginal distribution. We apply these three methods to DeepSeek-Coder-6.7B~\cite{deepseek-coder} and CodeLlama-7B~\cite{codellama}.

\paragraph{Implementation Details}
For GiFT, we repeat the description-to-code and code-to-description process for 20 times, and we generate 3 codes from each description and only select at most one correct code for the next round. To ensure fair generation times, for RFT, we generate 20$\times$3 codes for each seed description. And for RFT+RD, we rewrite the seed description into 5 new descriptions and generate 10 codes for each new description and the original description. For all the generation processes of LLMs, we set a temperature of 1.0. We set the temperature $T=2$ in weights calculation for random sampling. For three methods, we select $K=8$ codes per description for fine-tuning and employ resampling for descriptions with fewer than 8 codes. More details can be found in Appendix~\ref{app:implementation}.

\section{Experiments}
\label{sec:exp}
\paragraph{Overall Results}
We apply RFT, RFT+RD, and GiFT to DeepSeek-Coder-6.7B and CodeLlama-7B across four datasets with a 3-iteration self-training, the results are shown in Figure~\ref{overall_result}. Note that for RFT+RD and GiFT, we also consider codes generated from RFT as candidates for being selected, since we find that incorporating codes from RFT could further boost the performance of GiFT, even though merely using GiFT data has already outperformed baseline methods. We discuss the impact of these RFT-generated codes in Appendix~\ref{app:rft_in_gift}.

We could see that GiFT outperforms RFT and RFT+RD with a significant margin on all evaluated datasets, which indicates the effectiveness of GiFT. Generally, the improvement brought by GiFT is more significant on more challenging datasets like APPS+. We think this is because LLM's output distribution for complicated descriptions is more peaked, which exacerbates the bias in loss calculation.  This speculation is supported by the results shown at the bottom of Figure~\ref{fig:ppl_result}. We can find that the perplexity distribution of self-generated codes of APPS+ Introductory is much more peaked compared to that of MBPP+.

Given the fact that GiFT is superior compared to RFT, and RFT+RD outperforms RFT on most of the datasets, we demonstrate that drawing self-generated codes based on multiple possible descriptions that represent the intention of the seed description is better than drawing solely based on the seed description. In other words, mitigating the bias introduced in the loss calculation of examples from conditional distribution is beneficial for LLM fine-tuning.

\paragraph{Analysis for RFT+RD}
Though RFT+RD outperforms RFT on most of the datasets, drawing codes based on multiple rewritten descriptions is still not comparable with GiFT.

We find the reason is that the rewriting ability of LLMs is not satisfiable. There are often errors and information loss in rewritten descriptions, which makes the codes translated from rewritten descriptions often wrong. We calculate the pass rate of self-generated codes in the first iteration to indicate the correctness of self-generated descriptions.
For RFT+RD, we separately calculate the pass rate of codes from the seed description and five rewritten descriptions. The results are shown in Table~\ref{tab:sum_correct}.
We can see that the pass rate of codes generated from rewritten descriptions is significantly lower, which indicates that the rewritten descriptions are often incorrect. Since we do not train LLMs to rewrite descriptions, this phenomenon is expected to exist in the following iterations of self-training.
Given this reason, we believe that there will be no significant improvement despite scaling the RFT+RD method with more rewritten descriptions.

\begin{table}[h]
\small
\centering
\resizebox{0.9\columnwidth}{!}{%
\begin{tabular}{@{}lccc@{}}
\toprule
\multicolumn{1}{l}{\multirow{2}{*}{Datasets}} & \multicolumn{2}{c}{RD}                                               & \multicolumn{1}{c}{\multirow{2}{*}{GiFT}} \\ \cmidrule(lr){2-3}
                        & \multicolumn{1}{c}{Seed} & \multicolumn{1}{c}{Rewritten} &                     \\ \midrule
APPS+ (Intro.)                                        & 17.79       & 4.8              & >10.31                  \\
APPS+ (Inter.)                                       & 3.22        & 0.38             & >2.28                   \\
MBPP+                                         & 53.71                        & 24.6                              & >24.07                                  \\
CodeInsight                                   & 43.87                        & 9.84                              & >24.18                                   \\ \bottomrule
\end{tabular}
}
\caption{Pass rate (\%) of self-generated codes from the seed description, rewritten description, and GiFT. In GiFT, we generate 3 codes per description and save at most 1 for the next round. Thus, the true pass rate of GiFT is higher than the value in this table.}
\label{tab:sum_correct}
\vspace{-5pt}
\end{table}

\begin{figure}[h]
\centering
\includegraphics[width=0.99\columnwidth, trim=0 0 0 0, clip]{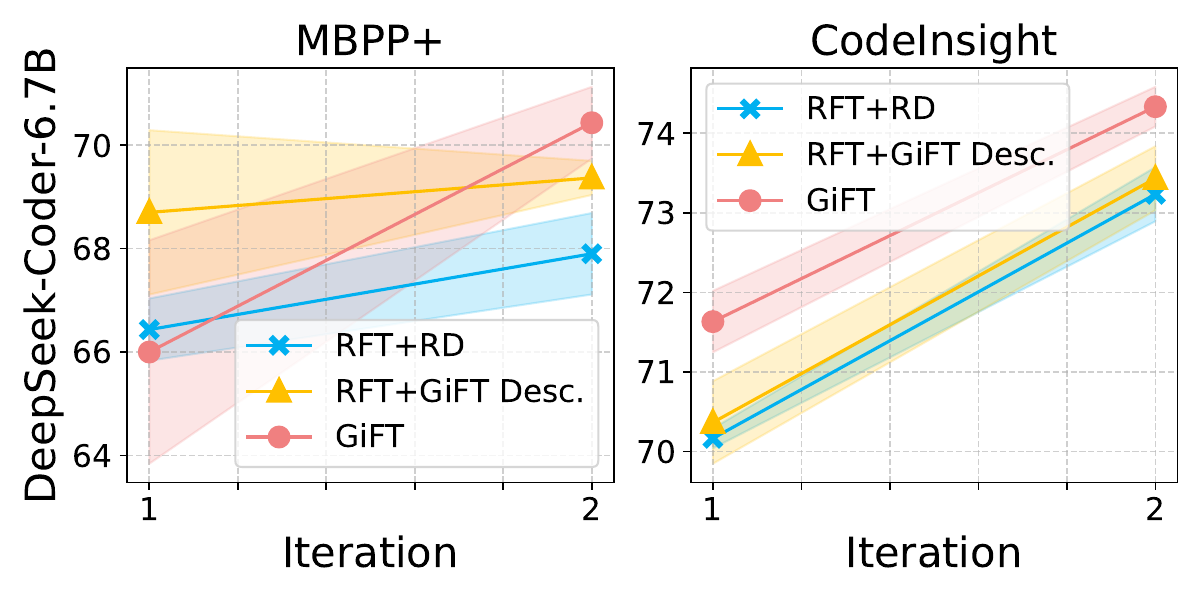}
\caption{Pass@1 (\%) of applying RFT+RD with descriptions from a single step of Gibbs sampling to Deepseek-Coder-6.7B on MBPP+ and CodeInsight.}
\label{fig:rft+gift_desc}
\vspace{-5pt}
\end{figure}

\paragraph{RFT+RD with Descriptions from a Single-step GiFT}

Given that the rewritten descriptions are often incorrect, we evaluate the performance of another stronger setting for RFT+RD. That is, we use the self-generated descriptions from a single step of Gibbs sampling as the rewritten descriptions in RFT+RD. The results of DeepSeek-Coder-6.7B on MBPP+ and CodeInsight are shown in Figure~\ref{fig:rft+gift_desc}. The improvement of stronger RFT+RD over the vanilla RFT+RD verifies our claim that it is better to leverage the outstanding ability of LLMs to translate the code back to a description instead of rewriting the description to a new one. However, we could see that GiFT still outperforms the stronger RFT+RD. We think this is because all of the descriptions are from the same code. Instead, in GiFT, descriptions are from various codes in each Gibbs sampling step.

\begin{figure}[h]
\centering
\includegraphics[width=0.99\columnwidth, trim=0 0 0 0, clip]{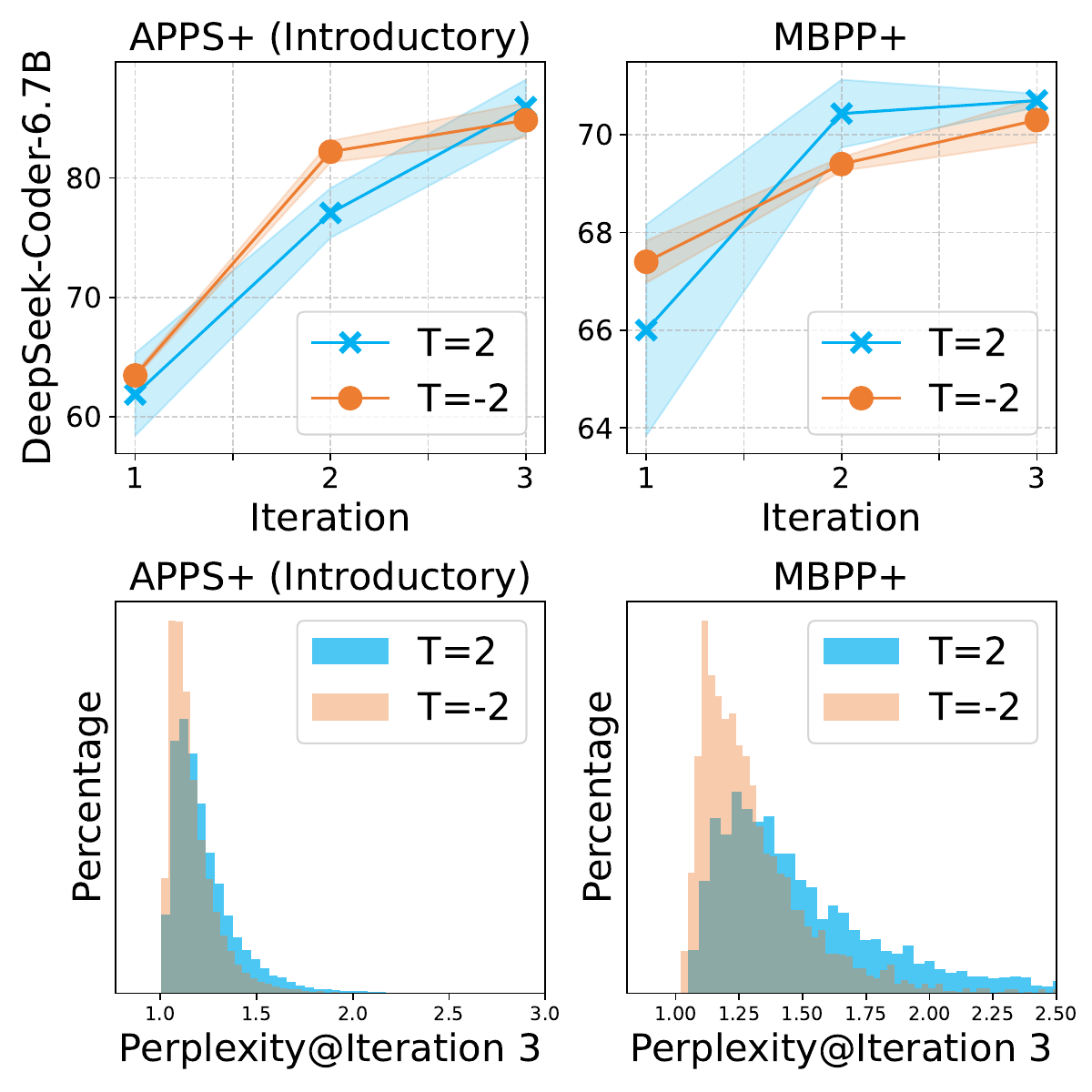}
\caption{\textbf{Top:} Pass@1 (\%) of applying GiFT to Deepseek-Coder-6.7B on APPS+ (Introductory) and MBPP+ with $T=\pm 2$. \textbf{Bottom:} Perplexity distribution of self-generated codes at the 3rd iteration for APPS+ (Introductory) and MBPP+.}
\label{fig:ppl_result}
\vspace{-5pt}
\end{figure}

\paragraph{Impact of $T$ in Data Selection}
Recall that in perplexity-guided data selection, we set $T=2$ to encourage the selection of more codes from the tails of the distribution to mitigate the tail narrowing problem~\cite{ding2024mitigating}. On the contrary, we could set $T$ as a negative value to select more codes from the head. By setting $T=\pm 2$, we explore the impact of data source (head or tail) for LLM fine-tuning. Note that we conduct extended experiments by setting $T=\pm 5$ in Appendix~\ref{app:addtional_T}.

We show the performance of DeepSeek-Coder-6.7B on APPS+ Introductory and MBPP+ in Figure~\ref{fig:ppl_result}. Furthermore, we visualize the perplexity distribution of self-generated codes at the third iteration for APPS+ Introductory and MBPP+. We could find that selecting more codes from the head outperforms the tail at the first several iterations, but is surpassed from the third iteration. We speculate that selecting more codes from the head reinforces LLM's knowledge at the head hence accelerating training at the beginning, but with the expense of discarding or forgetting knowledge at the tail. Over iterations, the tail-narrowing phenomenon is exacerbated and hinders further improvement. When selecting more codes from the tail, LLMs could achieve an overall better performance though they improve slower. As we show at the bottom of Figure~\ref{fig:ppl_result}, after three iterations, LLM could generate more low-perplexity codes on two datasets if we set $T=2$.

\begin{figure}[h]
\centering
\includegraphics[width=0.99\columnwidth, trim=0 0 0 0, clip]{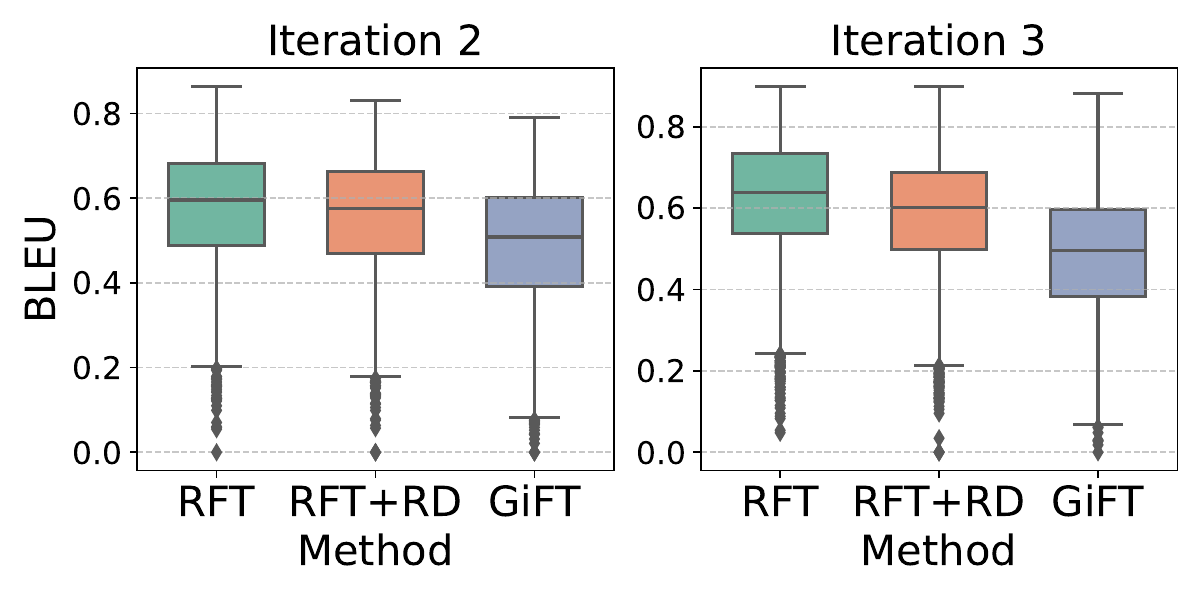}
\caption{Boxplot of BLEU for self-generated codes from DeepSeek-Coder-6.7B on MBPP+ at the 2nd and 3rd iteration.}
\label{fig:code_bleu}
\vspace{-5pt}
\end{figure}

\paragraph{Similarity Analysis for Self-generated Code}
We use BLEU as a measurement of code diversity to show that one of the benefits of using GiFT is the increase in data diversity. For each seed description, we calculate the BLEU score between any two self-generated codes and average them to indicate one code's similarity to others. We show the BLEU results of MBPP+ at the 2nd and 3rd iteration of DeepSeek-Coder-6.7B in Figure~\ref{fig:code_bleu}. We can observe that generating codes from various descriptions leads to more diverse codes for fine-tuning. Besides, as the number of iterations increases, RFT tends to generate more similar code, while the diversity holds for GiFT.

\begin{figure}[h]
\centering
\includegraphics[width=0.99\columnwidth, trim=0 0 0 0, clip]{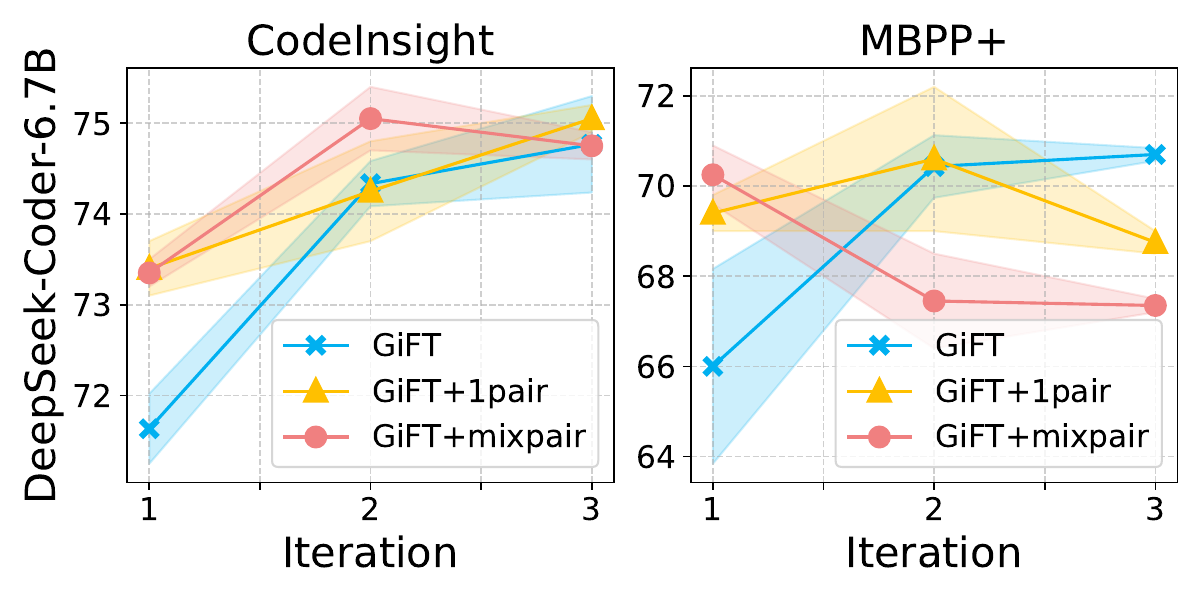}
\caption{Comparison of incorporating self-generated descriptions and vanilla GiFT on DeepSeek-Coder-6.7B over CodeInsight and MBPP+.}
\label{fig:use_sum}
\vspace{-5pt}
\end{figure}

\paragraph{Incorporating Self-generated Descriptions into Fine-tuning}
In GiFT, we only take the self-generated codes for fine-tuning after Gibbs sampling. Here we investigate the impact of incorporating self-generated descriptions into fine-tuning. Theoretically, if self-generated descriptions can match self-generated codes, LLMs are expected to achieve an even better performance, since LLMs benefit from not only diverse codes but also diverse descriptions as inputs.

We discover two alternatives, for each seed description, we add 8 self-generated descriptions, and 1) each one is paired with the code generated from it (denoted as GiFT-1pair). 2) each one is paired with 8 codes randomly sampled from the self-generated code set of the seed description (denoted as GiFT-mixpair). Note that not all self-generated descriptions are correct. We only select self-generated descriptions that can result in correct codes. 
We compare these two settings with the vanilla GiFT on DeepSeek-Coder-6.7B over CodeInsight and MBPP+ and the results are shown in Figure~\ref{fig:use_sum}. It was observed that incorporating self-generated descriptions into fine-tuning leads to better performance at the first iteration, yet is outperformed by GiFT in subsequent iterations.

We suspect that this is because LLMs are relatively tolerant of noisy data at the beginning, but as they have more expertise, their requirements for data quality become increasingly higher. We find there are mainly two sources of noisy pairs. First, some self-generated descriptions are just incorrect. Since we additionally provide some test cases in the docstring, LLMs may generate correct codes by inferring through given test cases and ignoring the incorrect description. Second, a self-generated description may not match all codes in the self-generated code set, possibly due to specifications on implementation requirements.
To filter out noisy pairs, we may calculate the similarity between description and code using code search models~\cite{li-etal-2023-rethinking-negative, li-etal-2022-exploring-representation} or LLM-as-a-Judge~\cite{llmasajudge, llmasajudgesurvey}. Since we mainly aim to show the benefit of fine-tuning LLMs using the seed description paired with codes from the marginal distribution, we leave this as our future work.

\begin{figure}[h]
\centering
\includegraphics[width=0.99\columnwidth, trim=0 0 0 0, clip]{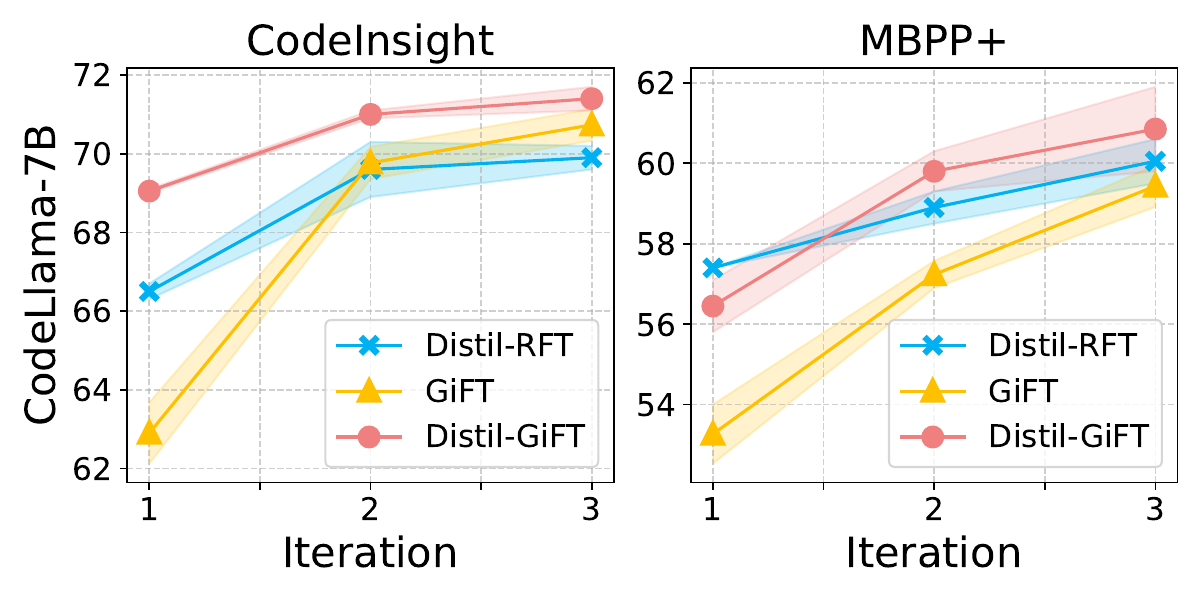}
\caption{Comparison of using GiFT and RFT data from DeepSeek-Coder-6.7B to distill CodeLlama-7B over CodeInsight and MBPP+. The yellow line shows the performance of self-training with GiFT.}
\label{fig:distil}
\vspace{-10pt}
\end{figure}

\paragraph{GiFT for Distillation}
As we discussed in the introduction, since the prerequisite of stronger LLMs limits the generalizability of distillation methods, we focus on self-training methods, as many recent works do. Yet, we are also interested in exploring whether drawing codes from the marginal distribution is beneficial for distillation. To simulate the distillation process, we use self-generated codes from DeepSeek-Coder-6.7B to fine-tune CodeLlama-7B, as we find that DeepSeek-Coder is stronger than CodeLlama on evaluated datasets. We keep other settings the same as they are in the main experiments. The comparison between distillation with RFT and GiFT on MBPP+ and CodeInsight is shown in Figure~\ref{fig:distil}. It is observed that GiFT also outperforms RFT under the distillation setting. This superiority meets our expectations because the benefit of fine-tuning LLMs with codes drawing from the marginal distribution is not limited to self-training methods, as we analyzed in the theoretical insights.

\begin{figure}[h]
\centering
\includegraphics[width=0.99\columnwidth, trim=0 0 0 0, clip]{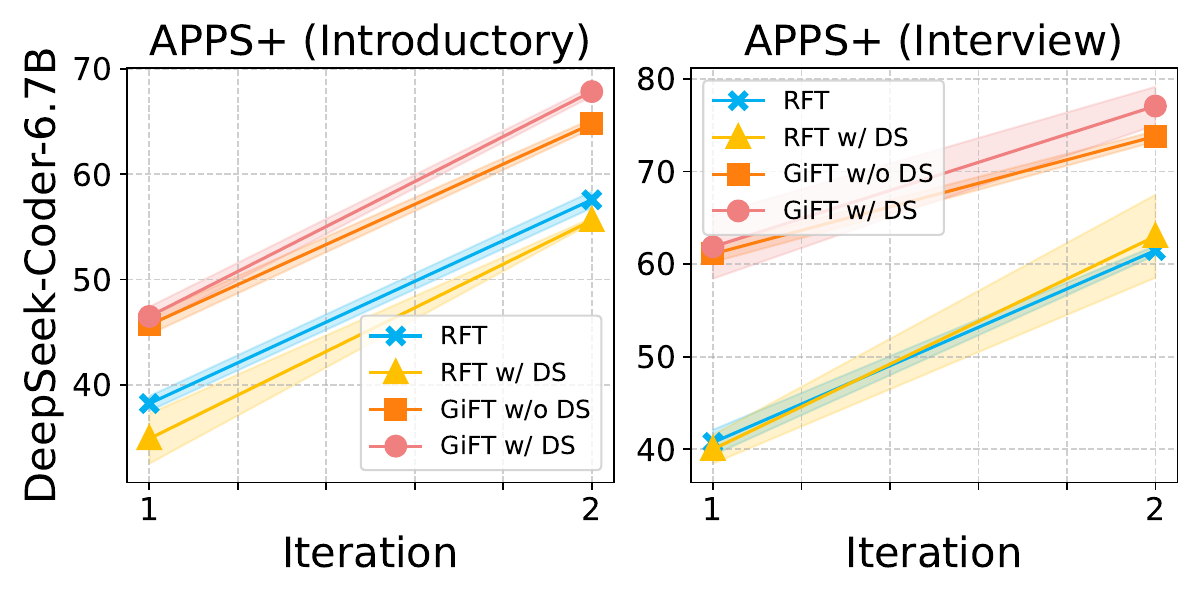}
\caption{Pass@1 (\%) of applying the proposed data selection method to RFT and GiFT on APPS+ (Introductory) and APPS+ (Interview). DS is short for data selection.}
\label{fig:rft+ds}
\vspace{-10pt}
\end{figure}

\paragraph{Ablation study for Perplexity-guided Data Selection}
The proposed perplexity-guided data selection method can be applied not only to GiFT but also to baseline methods. The results of applying the proposed data selection method to RFT and GiFT with $T=2$ on DeepSeek-Coder-6.7B are shown in Figure~\ref{fig:rft+ds}. We can see that the data selection method does not bring significant improvement to RFT compared with GiFT.

\begin{figure}[h]
\centering
\includegraphics[width=0.99\columnwidth, trim=0 0 0 0, clip]{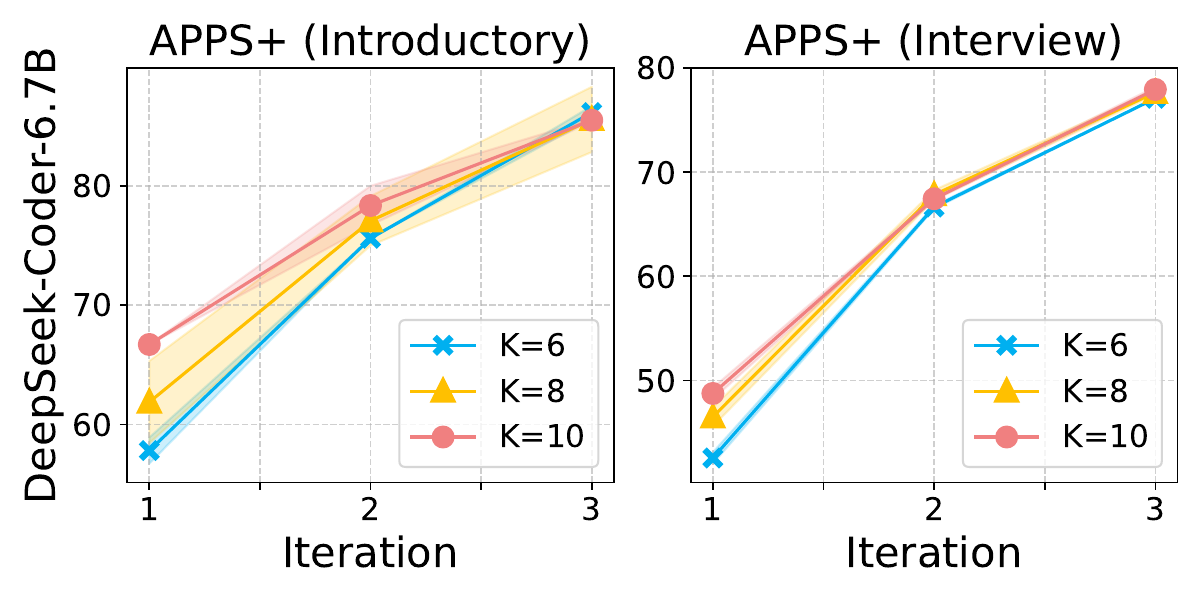}
\caption{Pass@1 (\%) of applying GiFT to DeepSeek-Coder-6.7B on APPS+ (Introductory) and APPS+ (Interview) with $K=6,8,10$, respectively.}
\label{fig:k_ablation}
\vspace{-10pt}
\end{figure}

\paragraph{Impact of $K$ in Data Selection}
In this paper, we set $K=8$ for all the reported experimental results. In this section, we explore the impact of $K$ in GiFT. We set $K=6,8,10$ for DeepSeek-Coder-6.7B on APPS+ Introductory and APPS+ Interview and the results are shown in Figure~\ref{fig:k_ablation}. We can find that pairing each seed description with more codes significantly improves LLM performance at the beginning of iterative self-training. Yet this benefit diminishes as the iteration progresses, and finally, LLM performance converges to similar performance. We think that the curves will converge to the upper bound of LLM's potential. Thus, we argue that increasing $K$ in GiFT accelerates iterative self-training.

\section{Discussion}
Here we discuss the generalization of GiFT to other tasks. The applicability of GiFT and its superiority over RFT depends on two factors. 
First, there is a joint input-output sampling space, in which same intention has multiple possible forms of presentation, and such presentation largely decides the self-generated outputs from LLM. GiFT is suitable for code generation because descriptions and codes naturally form such a joint space~\cite{reco}. On the other hand, take question answering as an example, while there are numerous ways to phrase the same question, the answers tend to be highly similar due to the uniqueness of factual truths. This characteristic naturally mitigates biases introduced by conditional sampling.

Second, LLMs should be able to translate accurately between inputs and outputs. Since LLMs are found to be good at translating between descriptions and codes~\cite{codesumm, nl2code:survey,codegensurvey}, GiFT performs well on code generation.
Nevertheless, for mathematical reasoning, whether LLMs can reliably generate a math problem based on the given solution should be carefully evaluated before we apply GiFT. If the translation from solutions to problems lacks precision, the Gibbs sampling process in GiFT may be highly inefficient.

\section{Conclusion}
In this paper, we first theoretically demonstrate 
the benefit of fine-tuning LLMs with codes from the marginal distribution of the joint description-code space instead of the conditional distribution conditioned on the seed description. Then, we propose GiFT, which iteratively translates natural language descriptions and codes between each other to approximate the marginal distribution. Furthermore, we leverage perplexity to guide data selection to mitigate the data imbalance problem in Gibbs sampling. Experimental results on two LLMs across four datasets demonstrate the effectiveness of GiFT.

\section*{Acknowledgement}
We thank the anonymous reviewers for their helpful
comments and suggestions. This research is supported by the RIE2025 Industry Alignment Fund – Industry Collaboration Projects (IAF-ICP) (Award I2301E0026), administered by A*STAR, as well as supported by Alibaba Group and NTU Singapore through Alibaba-NTU Global e-Sustainability CorpLab (ANGEL).

\section*{Limitations}
There are mainly three limitations in this work. First, GiFT is only evaluated on LLMs with a size of around 7B across Python datasets. However, as we demonstrated in the theoretical analysis, the loss bias from the conditional distribution is independent of model sizes and programming languages. Thus, we expect that GiFT is also effective in bigger or smaller LLMs and other programming languages. Second, GiFT relies heavily on test cases to filter out wrong self-generated codes. In this paper, we mainly evaluate GiFT on datasets that already provide test cases in the training set. We do not evaluate GiFT on the most recent high-quality datasets like OSS-Instruct~\cite{magicoder}. A possible solution is to ask LLMs to generate test cases and codes at the same time, which is studied by recent works~\cite{codet,b4,dstc}. Third, GiFT outperforms RFT by translating between natural language descriptions and codes, which introduces additional code summarization steps in the self-generation process. In our future work, we will focus on improving the sampling efficiency of GiFT by introducing a dynamic sampling strategy. For example, we do not generate a fixed number of codes and select at most one correct code for the next round. Instead, we dynamically adjust the number of generated codes based on the estimated accuracy to ensure that at least one correct code is expected. Thus, LLMs tend to generate fewer code candidates for easier descriptions in each round.

\bibliography{custom}

\appendix

\section{Prompting Templates for Code-to-Text and Text-to-Code Translation}
\label{app:prompt}
An example of a prompt for code generation is shown in Table~\ref{tab:codegen_prompt}. We construct a template for code completion where we provide a function head and a function docstring. The function docstring could be replaced with descriptions from previous Gibbs sampling results. We additionally provide one input-output pair for MBPP-sanitized and CodeInsight.

An example of a prompt for code summarization is shown in Table~\ref{tab:sumgen_prompt}. We provide in-context examples to help LLM learn summarization. The in-context example is selected from a pool. We construct the pool by asking LLMs to summarize the same dataset without in-context examples and then filter out meaningless or too-long responses. The code following ``\verb|###|Code:'' could be replaced with codes from previous Gibbs sampling results.

\begin{table}[h]
\begin{tabular}{p{0.9\columnwidth}}
\toprule
def first\_repeated\_char(str1): \\
\hspace{15pt}    """ Write a python function to find the first repeated character in a given string. \\
\hspace{15pt}    \verb|>>>| first\_repeated\_char("abcabc")\\
\hspace{15pt}    "a" \\
\hspace{15pt}    """ \\
\bottomrule
\end{tabular} 
\caption{A prompt example of MBPP-sanitized/1 for code generation.}
\label{tab:codegen_prompt}
\end{table}

\begin{table}[h]
\begin{tabular}{p{0.9\columnwidth}}
\toprule
\verb|###|Code: \\
\{example\_code\} \\
\verb|###|Description of the given code: \\
\{example\_description\} \\
\\
\\
\verb|###|Code: \\
def first\_repeated\_char(str1):\\
\hspace{15pt}     \verb|###| BEGIN SOLUTION \\
\hspace{15pt}     letters\_found = [] \\
\\
\hspace{15pt}     for char in str1: \\
\hspace{15pt} \hspace{15pt}         if char in letters\_found: \\
\hspace{15pt} \hspace{15pt} \hspace{15pt}             return char \\
\hspace{15pt} \hspace{15pt}         else: \\
\hspace{15pt} \hspace{15pt} \hspace{15pt}             letters\_found.append(char) \\
\hspace{15pt}     \verb|###| END SOLUTION \\
\\
\verb|###|Description of the given code: \\
\bottomrule
\end{tabular} 
\caption{A prompt example used of MBPP-sanitized/1 for code summarization. \{example\_code\} and \{example\_description\} are randomly selected from a pool.}
\label{tab:sumgen_prompt}
\end{table}

\begin{table}[h]
\begin{tabular}{p{0.9\columnwidth}}
\toprule
Rewrite the given Description \\
\verb|###|Description: \\
Write a python function to find the first repeated character in a given string. \\
\verb|###|New Description: \\
\bottomrule
\end{tabular} 
\caption{A prompt example of MBPP-sanitized/1 for rewriting description.}
\label{tab:rewrite_prompt}
\end{table}

\section{Algorithm Workflow}
\label{app:algo}
The algorithm workflow of GiFT in each iteration is shown in Algorithm~\ref{algo}. This workflow is repeated where the updated LLM $\mathcal{M}^*$ in each iteration is used as the initialization LLM for the next one.

\section{Experimental Setup}
\subsection{Dataset Statistics}
\label{app:setup}
The dataset statistics are shown in Table~\ref{tab:dataset_stat}. Note that we use MBPP sanitized version for training while MBPP+ (a version with more test cases for each problem) for testing.

\begin{table}[h]
\centering
\resizebox{\columnwidth}{!}{%
\begin{tabular}{@{}lcccc@{}}
\toprule
\multirow{2}{*}{Dataset} & APPS+ & APPS+  & \multirow{2}{*}{MBPP} & \multirow{2}{*}{CodeInsight} \\ 
        & Introductory & Interview &  &  \\ \midrule
Train   & 1,998   & 3,736  & 170 & 1,547  \\
Test    & 90   & 367  & 378 & 1,856  \\ \bottomrule
\end{tabular}
}
\caption{Statistics of the dataset used in our experiment.}
\label{tab:dataset_stat}
\end{table}

\begin{figure*}[ht]
\centering
\includegraphics[width=1.99\columnwidth, trim=0 0 0 0, clip]{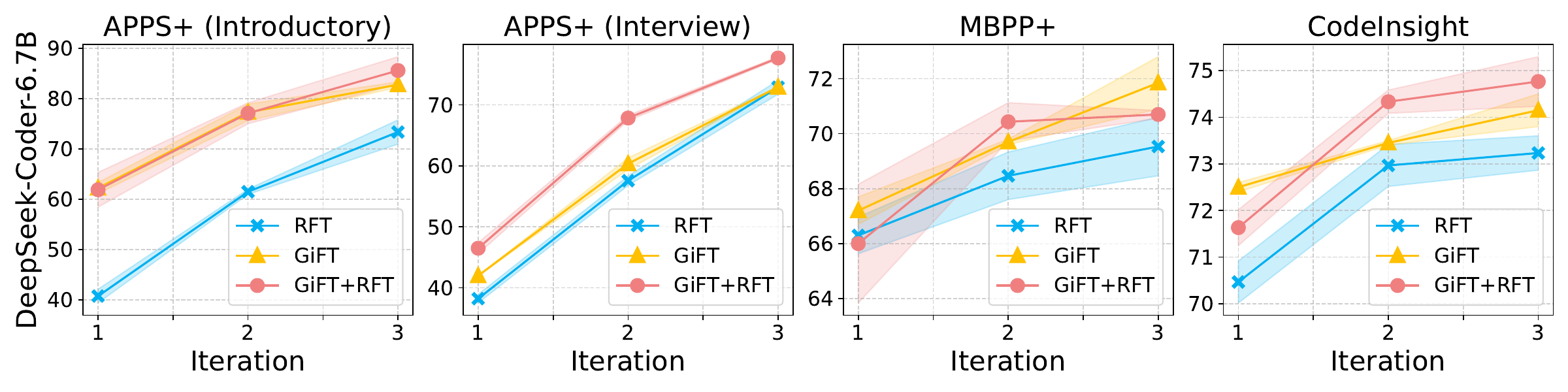}
\caption{Pass@1 (\%) of applying RFT, pure GiFT, and GiFT+RFT to Deepseek-Coder-6.7B across four code generation datasets. The x-axis represents the iteration number and the shaded area represents the standard deviation.}
\label{fig:impact_of_rft}
\end{figure*}

\subsection{Implementation Details}
\label{app:implementation}
For the prompting template of generating codes in RFT and RFT+RD, we follow Table~\ref{tab:codegen_prompt}. An example of a prompt for rewriting descriptions is shown in Table~\ref{tab:rewrite_prompt}. DeepSeek-Coder-6.7B is initialized with the checkpoint at \url{https://huggingface.co/deepseek-ai/deepseek-coder-6.7b-base}. CodeLlama-7B is initialized with the checkpoint at \url{https://huggingface.co/meta-llama/CodeLlama-7b-hf}. We fine-tune the LLMs with DeepSpeed ZeRO-2 optimization with a batch size of 1 for each GPU. The maximum length is set to be 1,024 for MBPP and Code Insight and 1,536 for APPS. We use AdamW as the optimizer with a learning rate of 2e-5 and set the gradient accumulation steps as 16. We fine-tune LLMs for 2 epochs for APPS+ and 1 epoch for MBPP and CodeInsight. All experiments are running with 3 random seeds 1234, 12345, and 123456. Experiments are conducted on 8 Nvidia Tesla A100 GPUs.

\section{More Experimental Results}

\subsection{Impact of RFT data in GiFT}
\label{app:rft_in_gift}
As we mentioned, we find that incorporating RFT data in GiFT could further boost the performance of GiFT, even though merely fine-tuning LLMs with codes from GiFT has already outperformed RFT. We compare the performance of RFT, pure GiFT, and GiFT+RFT on DeepSeek-Coder-6.7B across four datasets in Figure~\ref{fig:impact_of_rft}. We suspect that this is due to the lack of correct codes for some seed descriptions. As a result, we have to resample existing codes to ensure that there are $K$ codes for each seed description during fine-tuning. The incorporation of RFT data mitigates this drawback and improves data diversity. In practice, we can ask LLMs to generate several times at the first round of GiFT, since in the first round of GiFT, LLMs generate codes from the seed description, which is the same in input of RFT.

\begin{figure}[h]
\centering
\includegraphics[width=0.99\columnwidth, trim=0 0 0 0, clip]{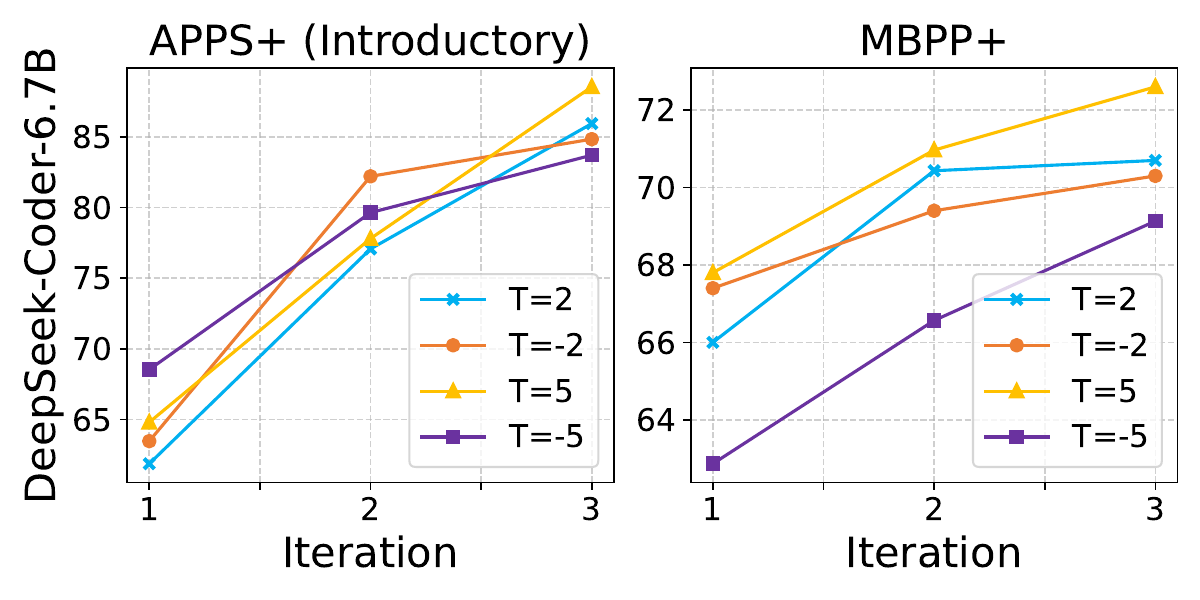}
\caption{Pass@1 (\%) of applying GiFT to Deepseek-Coder-6.7B on APPS+ (Introductory) and MBPP+ with $T=\pm 2,\pm 5$.}
\label{fig:additional_T}
\end{figure}

\subsection{Extended Experiments of $T$}
\label{app:addtional_T}
We additionally conduct experiments with bigger and smaller $T$ values to further study the impact of $T$ in data selection. Specifically, we set $T=\pm 5$ and the results are shown in Figure~\ref{fig:additional_T}. We can observe that a larger $T$ which makes codes from the tail more likely to be selected leads to even better performance. On the contrary, $T=-5$ leads to worse performance. We argue that $T$ should be set within a moderate range and specifically tuned for each dataset.

\begin{algorithm*}[t]
\caption{Workflow of GiFT in each iteration}
\label{alg:gibbs_sampling}
\textbf{Input}: A seed dataset $\mathcal{D} = \{d_i\}_{i=1}^{N}$, an LLM $\mathcal{M}$. \\
\textbf{Parameter}: Gibbs sampling iterations $n$, selection threshold $K$, temperature $T$. \\
\textbf{Output}: An updated LLM $\mathcal{M}^*$ for next GiFT iteration.

\begin{algorithmic}[1]

\State $\mathcal{C} \gets \emptyset$.

\For{each description $d_i \in \mathcal{D}$} \Comment{Gibbs Sampling}
    \State $c_{i1} \gets \mathcal{M}(d_i)$
    \For{$k \gets 1$ to $n$}
        \State $d_{ik} \gets \mathcal{M}(c_{ik})$ \Comment{Summarize code into description}
        \State $c_{ik+1} \gets \mathcal{M}(d_{ik})$ \Comment{Generate code from description}
    \EndFor
    \State $\mathcal{C}_i \gets \{c_{i1}, c_{i2}, \dots, c_{in}\}$
\EndFor

\State $\mathcal{C}^* \gets \emptyset$

\For{each description $d_i \in \mathcal{D}$} \Comment{Perplexity-Guided Selection}
    \For{each code $c_{ij} \in \mathcal{C}_i$}
        \State $\verb|ppl|(c_{ij}) \gets \exp \Big( -\frac{1}{|c_{ij}|} \sum_{t=1}^{|c_{ij}|} \log P_{\mathcal{M}}(c_t | c_{<t}, d_i) \Big)$ \Comment{Compute perplexity}
    \EndFor
    \For{each code $c_{ij} \in \mathcal{C}_i$}
    \State $w_{ij} \gets \frac{e^{\text{ppl}(c_{ij}) / T}}{\sum_{j=1}^{n_i} e^{\text{ppl}(c_{ij}) / T}}$ \Comment{Compute weight}
    \EndFor
    \State $\{c_{ij}\}^{K} \gets \verb|weighted random sampling|(\{c_{ij},w_{ij}\})$
    \State $\mathcal{C}_i^* \gets \mathcal{C}_i^* \cup \{c_{ij}\}^{K}$
\EndFor

\State $\mathcal{D}^* \gets \{(d_i, c) | d_i \in \mathcal{D}, c \in \mathcal{C}_i^*\}$ \Comment{Construct dataset for SFT}
\State $\mathcal{M}^* \gets SFT(\mathcal{M}, \mathcal{D}^*)$ \Comment{Supervised Fine-Tuning}

\State \Return $\mathcal{M}^*$ \Comment{Return fine-tuned model}

\end{algorithmic}
\label{algo}
\end{algorithm*}

\end{document}